\documentclass{egpubl}

\SpecialIssuePaper

\usepackage[T1]{fontenc}
\usepackage{dfadobe}  

\usepackage{cite}
\BibtexOrBiblatex
\electronicVersion
\PrintedOrElectronic
\ifpdf \usepackage[pdftex]{graphicx} \pdfcompresslevel=9
\else \usepackage[dvips]{graphicx} \fi

\usepackage{egweblnk} 

\usepackage{amssymb}
\usepackage{booktabs}
\usepackage{multirow}
\usepackage{tabularx}
\usepackage{enumitem}
\newcolumntype{Y}{>{\centering\arraybackslash}X}
\usepackage{adjustbox}

\newcommand{\bc}{\mathbf{c}}

\newcommand{\bn}{\mathbf{n}}

\newcommand{\bp}{\mathbf{p}}

\newcommand{\bu}{\mathbf{u}}
\newcommand{\bv}{\mathbf{v}}

\newcommand{\bx}{\mathbf{x}}
\newcommand{\by}{\mathbf{y}}

\newcommand{\bP}{\mathbf{P}}

\newcommand{\bR}{\mathbf{R}}

\newcommand{\mD}{\mathcal{D}}
\newcommand{\mP}{\mathcal{P}}

\newcommand{\mN}{\mathcal{N}}

\title[Learning Part Boundaries from 3D Point Clouds]%
      {Learning Part Boundaries from 3D Point Clouds
      \vspace{-5mm}
      }

\author[Marios Loizou, Melinos Averkiou, Evangelos Kalogerakis]
{\parbox{\textwidth}{\centering Marios Loizou $^{1,2}$, Melinos Averkiou $^{1,2}$, Evangelos Kalogerakis$^3$}
\vspace{-2mm}
        \\
{\parbox{\textwidth}{\centering $^1$ University of Cyprus, 
         $^2$ RISE CoE, Nicosia, Cyprus,
         $^3$  	University of Massachusetts Amherst
       }
}
}

\newcommand\blfootnote[1]{%
  \begingroup
  \renewcommand\thefootnote{}\footnote{#1}%
  \addtocounter{footnote}{-1}%
  \endgroup
}

\begin{document}

\pagestyle{empty}
\maketitle
\thispagestyle{empty}

\begin{abstract}
We present a method that detects  boundaries of parts in 3D\ shapes represented as point clouds. Our method is based on a graph convolutional network architecture that outputs a probability for a point to lie in an area that separates two or more parts in a 3D\ shape.
Our boundary detector is quite generic: it can be trained to  localize boundaries of  semantic parts or geometric primitives commonly used in 3D\ modeling. Our experiments demonstrate that our method can extract more accurate boundaries that are closer to ground-truth ones  compared to alternatives. We also demonstrate an application of our network to fine-grained semantic shape segmentation, where we also show improvements in terms of part labeling performance.



\end{abstract}

\blfootnote{This is the author's version of the work. It is posted here for your personal use. The definitive version of the article will be published at Computer Graphics Forum, vol.39, no.5, 2020, https://doi.org/10.1111/cgf.14078}


\begin{figure*}[tbp]
     \centering
     \includegraphics[width=1\textwidth]{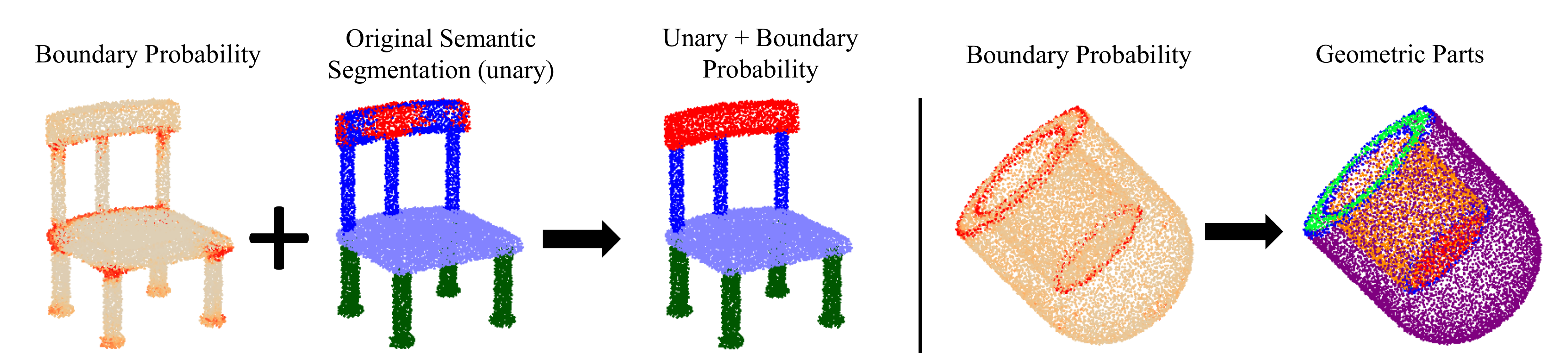}
     \caption{Our method predicts part boundaries in 3D point clouds using a graph convolutional network which outputs a probability per point to lie on a boundary between parts in a 3D shape. The output probability per point can be used in pairwise terms to improve graph-based semantic segmentation methods (left) by localizing boundaries between semantic parts. It can also be used in the geometric decomposition of point clouds into regions enclosed by sharp boundaries detected by our method (right). }
     \label{fig:teaser}
     \vspace{-5mm}
 \end{figure*}

\section{Introduction}

Segmenting 3D objects into their constituent parts with accurate boundaries is a fundamental problem in computer graphics and vision. Although there has been significant amount of research  in  detecting contours and  object boundaries in natural images
with neural networks \cite{Bertasius2015,BertasiusST15,Bertasius2016,castrejon2017annotating,hou2018birds,Kokkinos2015PushingTB,le2019rednet,Linsley2018,liu2018semantic,Liu2016,Man:17:cob,Wei2015,Wang2018,Saining2017}, detecting   boundaries in 3D\ point clouds is  largely an unexplored area. Despite  the significant advances in the area of 3D\ deep learning for processing unstructured point clouds,  most research has focused so far on assigning part tags  to individual points. The resulting segmentations often suffer from artifacts  at areas that lie near the boundaries of parts, since the point assignments become highly uncertain at these areas   (see also Figure \ref{fig:teaser}).

In this paper, we present a neural network approach that learns to detect part boundaries in point clouds of 3D\ shapes. There is a number of technical challenges to overcome in developing an approach that addresses this problem. First, the notion of an object part is often ambiguous and usually depends on the task. For example, in semantic segmentation, parts follow label definitions (e.g., leg, back, seat for chairs), while for 3D modeling tasks, shapes are often modeled as collections of geometric primitives  (e.g., spheres, cylinders, surfaces of extrusion, NURBS, and so on). We show that an effective boundary detector can be trained
from semantic segmentation datasets to accurately  localize boundaries of  labeled parts, and also from shape datasets with segmented geometric patches. Second, boundaries are usually sparse; only a small percentage of points in a point cloud lie near  boundaries. During training, we employ a sampling procedure to gather a sufficient amount of boundary points for training, and use a classification loss function robust to the imbalance of the number of boundary versus non-boundary points. Furthermore, in contrast to semantic segmentation networks that often rely on points expressed in global coordinate frames, we found that learning features from points expressed in local frames aligned with surface normals are better suited for boundary extraction. The output of our method is probabilistic: it assigns a probability for each point belonging to a part boundary or not. We demonstrate pairwise terms
that can easily adopt these probabilities within graph cuts formulations.  

We conducted a number of experiments to validate our method. First, we compare our extracted boundaries with  annotated ones in geometric and semantic segmentation tasks. 
We found that the boundaries produced by our architecture are much closer to ground-truth ones compared to alternatives. For example, we observed that the error was reduced by  $61.2\%$ compared to the  best alternative edge detector we adapted for our task (EC-Net, \cite{yu2018ec}), measured based on Chamfer distance between detected and ground-truth boundaries in the ABC dataset \cite{Koch2019}.
We also show that our boundary detector, when combined with graph cuts, offers a small, but noticeable boost in terms
the semantic segmentation performance: an increase of $+2.6\%$ in shape Intersection over Union (IoU), and $+0.5\%$ in part IoU on average in PartNet \cite{Mo_2019_CVPR}
compared to using a neural network (DGCNN \cite{Wang19}) that assigns tags to points without explicitly considering boundaries. Our contributions can be summarized as follows:

\begin{itemize}[noitemsep,leftmargin=*]
\item a neural network module, called LocalEdgeConv (inspired by DGCNN), that operates on point cloud neighborhoods expressed in local frames (in constrast to global frames used in \cite{Wang19}). We found that this adaptation is more suitable for the task of 3D\ part boundary detection.
\item a network training procedure that robustly samples and weights boundary data of either semantic parts or geometric primitives.
\item a graph cuts formulation that uses our probabilistic boundary detector to improve semantic shape segmentation, especially near part boundaries.
\end{itemize}

\section{Related Work}

Below we briefly overview prior  work on 3D point cloud segmentation as well as contour detection in 2D\ natural images and 3D\ shapes. 

\paragraph*{3D point cloud segmentation.} A large number of works have been proposed to segment point clouds by assigning a part label for each point. Most recent semantic segmentation approaches train deep neural networks that learn representations from point clouds based on point set abstraction and aggregation functions 
\cite{Qi17,Jiaxin:sonet,srivastava2019geometric,Komarichev2019,Le:deeper:2019,sharma2019learning},
point convolution operators 
\cite{Li:pointcnn,Hua17,liu2019densepoint,accv2018Groh,Atzmon18,Hermosilla18,xu2018spidercnn,Wu18:pointconv,Thomas19:kpconv},
graph convolution \cite{Wang19,Liu2019,Li:deepgcn,jiang2019hierarchical,xu2019gridgcn,Wang2019gcn,petrov2020crossshape},
convolution on hierarchical grids \cite{Riegler2017OctNet,klokov2017escape,wang2017ocnn,Wang:2018:AOP,su18splatnet},
point-to-voxel mappings \cite{rethage2018eccv,Shi_2019_CVPR,Liu2019:pointvoxelcnn}, 
view-based projections 
\cite{Kalogerakis:2017:ShapePFCN,Huang:2017:LMVCNN},
and spectral approaches \cite{Yi2017:syncspec,Bronstein16}. Apart from semantic segmentation, a number of approaches have also been proposed to perform geometric decompositions of point clouds based on convexity analysis \cite{vankaick14convseg,Stein14,deng2019cvxnet}, 
 primitive fitting \cite{Lingxiao:SFPN,schnabel-2007-efficient,Yangyan11,zou20173dprnn,sharma2020parsenet}, graph cuts \cite{Golovinskiy:2009:MBS,Karpathy_ICRA2013,Kisner18}, and clustering \cite{Rusu07,bogoslavskyi16iros,Zhang:3dv:2019}. In both semantic segmentation and geometric decomposition scenarios, part boundaries often tend to become fuzzy and noisy
(Figure \ref{fig:teaser} - left). To improve the quality of segmentation and align boundaries with underlying surface feature curves, such as creases, some methods employ simple geometric criteria, most commonly normal differences \cite{pham-jsis3d-cvpr19,Karpathy_ICRA2013,Stein14}, within pairwise terms modeling the probability of boundaries between points. Similar pairwise terms have  also been used in mesh segmentation approaches
 \cite{Katz03,Golovinskiy08,Kalogerakis:2010:labelMeshes,vankaick14convseg}.

In contrast, our method learns a pairwise term indicating the existence of part boundaries directly on 3D point clouds. As we show in our experiments, our learned boundaries are more accurate and are able to improve the quality of semantic segmentation  and geometric decomposition  to a larger degree compared to other alternatives.        

\paragraph*{Feature curve and edge detection on 3D\ shapes.} Our work is also related to learning methods for detecting edges, or feature curves on 3D\ shapes. This is because part boundaries often coincide with surface feature curves, such as creases, ridges and valleys. Detecting such feature curves relies on geometric features, such as curvature extrema or normal discontinuities that can be detected on meshes, RGB-D data or point clouds \cite{ohtake2004ridge,Belyaev05,ChoiTC13,Cao2017,Kolomenkin09,ddlinedrawing09,Sunkel11,Huang13}, however, their extraction often depends on hand-tuned procedures. Most similar to our approach, EC-Net \cite{yu2018ec} attempts to learn edges on point clouds using a deep architecture operating on isolated point cloud regions (patches).  Our method instead trains a neural network that operates directly on the whole point cloud encoding both local and global structure  without any patch extraction or pre-processing. In our experiments, we show that our approach is much more accurate for part boundary detection compared to EC-Net even when the latter is trained on the same dataset as our method.

\paragraph*{Contour detection for natural images.} Our approach is inspired by  contour detection methods for object segmentation in natural images.
Most recent approaches achieve state-of-the-art performance on contour detection by training deep convolutional or recurrent neural networks
\cite{Kokkinos2015PushingTB,Bertasius2015,BertasiusST15,Bertasius2016,Wei2015,Liu2016,Man:17:cob,castrejon2017annotating,Saining2017,Wang2018,hou2018birds,liu2018semantic,Linsley2018,le2019rednet}.
In contrast to the regular 2D grid structure of images, point clouds are unorganized and non-uniformly sampled. Our method adapts neural networks for point cloud processing and is trained on datasets for 3D\ semantic or geometric segmentation of shapes.\

\section{Method}

\begin{figure*}[t!]
     \centering
     \includegraphics[width=1\textwidth]{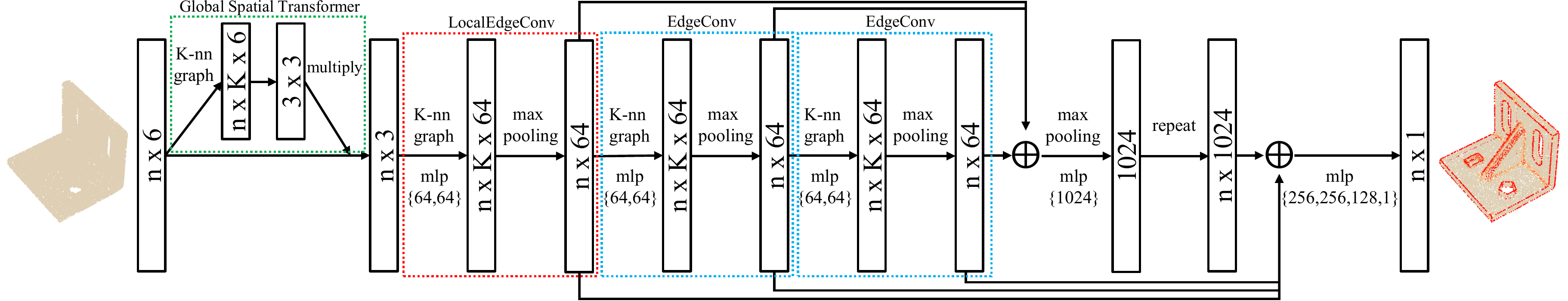}
     \caption{Architecture of our network (PB-DGCNN) for probabilistic boundary detection. It consists of three main blocks: the LocalEdgeConv layer, EdgeConv layers \cite{Wang19}, and a Global Spatial
     Transformer. The LocalEdgeConv layer constructs a $K$-NN graph for each input point $i$ in Euclidean space and expresses the coordinates of its $K$ nearest neighbors in the local coordinate frame at point $i$. Then a feature transformation is applied to the edge features of the graph through a 2-layer MLP, and output representations are aggregated through max-pooling. These representations are further processed by the two EdgeConv layers that operate on the $K$-nn graphs constructed in the feature space of the previous layer. Finally, a global descriptor is aggregated and concatenated with the point-wise descriptors of the three previous layers, which are transformed through a 4-layer MLP, and the boundary probability is produced by a sigmoid function. The Global Spatial Transformer \cite{Wang19} operates on the Euclidean space of the input points and helps to align the point cloud to a canonical space.}
     \label{fig:architecture}
\vspace{-5mm}     
 \end{figure*}

Our architecture takes as input a point cloud $\bP=\{\bp_i, \bn_i\}_{i=1}^N$, where $\bp_i$ are 3D point coordinates and $\bn_i$ are 3D\ normals, and outputs a scalar $b_i\in[0,1]$ for each point. The output $b_i$ represents the probability for
a part boundary to lie on the point $i$. Our architecture is shown in Figure \ref{fig:architecture}. We first describe it at test time (Section \ref{sec:test}), then we discuss the datasets used for training it, along with the training procedure (Sections \ref{sec:datasets} and \ref{sec:loss}). We present an application of our probabilistic boundary detector to semantic shape segmentation in Section \ref{sec:segmentation}.

\subsection{Architecture}
\label{sec:architecture}
\label{sec:test}

Our architecture, called PB-DGCNN, follows the concept of graph edge convolution (EdgeConv) introduced in the DGCNN network \cite{Wang19}. To implement edge convolution, a graph first needs to be formed over the point cloud. In the first EdgeConv layer, each point $i$ is connected to its $K$  neighbors in Euclidean space, where $K$ is a hyper-parameter of the network. In the original DGCNN formulation, the first EdgeConv layer processes the input point coordinates $\bp_i$ of each point $i$ along with the coordinates of its  neighbors $\{\bp_j\}_{j\in\mN_e(i)}$, where $\mN_e(i)$ is the Euclidean neighborhood of the point $i$. The output representation for each point is computed as follows: 
\begin{equation} 
\by_i = \max \limits_{j \in \mN_e(i)} MLP(\bp_i, \bp_j - \bp_i)
\label{eq:first_edgeconv}
\end{equation}
where $MLP$ represents a learned Multi-Layer Perceptron operating on the above input feature vector of point coordinates and differences (concatenated and flattened). In this manner, each edge encodes the input coordinates at a point $i$ along with the coordinates of its neighbors expressed relative to it.  The max operator (max pooling) is used to aggregate all edge representations per point and guarantees invariance to point and edge permutations. 

\paragraph*{LocalEdgeConv layer.} The relative point coordinate differences $(\bp_j - \bp_i)$ help capturing local neighborhood structure in the original DGCNN. However,  these differences are still expressed with respect to the global coordinate frame axes. As a result, if the local neighborhood is rotated, the input edge features to the MLP will change. In turn, the output of the MLP\ may also change. This effect might be desirable in the case of semantic segmentation, since changing the orientation of a part in a shape may change its functionality and its semantic label, especially for man-made objects (e.g. rotating a horizontal tailplane 90 degrees in an airplane would make it look like a vertical stabilizer). However, the part boundaries are more likely to remain unaffected by such local rotations e.g., one would still want to label points between the fuselage and the tailplane, or the  stabilizer, as boundaries. Thus, in our architecture, we make the following modification to the first edge convolution layer:    
\begin{equation} 
\by_i = \max \limits_{j \in \mN_e(i)} MLP \big( \bp_i, \bR_i^T(\bp_j - \bp_i) \big)
\label{eq:local_edgeconv}
\end{equation}
where $\bR_i$ is a  rotation matrix responsible for  expressing the relative point coordinate differences in a local coordinate frame at point $i$ (instead of a global one). Note that the transpose of the rotation is used to perform the coordinate transformation. The local frame is formed from the point normal $\bn_i$ and two tangent vectors 
$\bu,\bv$ randomly selected on the tangent plane of the point $i$:     
$\bR_i = [\bu_{i} \,\,\,  \bv_{i} \,\,\, \bn_i]$. Since the tangent vectors are chosen randomly, rotational invariance is not guaranteed, however, the use of the point normal decreases the variance of inputs that the network needs to handle. We call the above edge convolution of Eq. \ref{eq:local_edgeconv}\ as LocalEdgeConv in the rest of the paper. Experimentally, we observed a significant improvement in boundary detection due to  LocalEdgeConv (see our evaluation in Section \ref{sec:eval}).  We note that we also experimented with using principal curvature directions as tangent directions, and also treating their sign ambiguity through max pooling, yet the gain was still smaller than LocalEdgeConv (see results section). 

\paragraph*{LocalEdgeConv layer with normals as features.} Another variant of LocalEdgeConv we experimented with was to include point normals as additional input features to this layer. Specifically, we horizontally concatenate point positions and normals per point ($\bx_i = [\bp_i, \bn_i]$), then transform them through a MLP in a local coordinate system:
\begin{equation} 
\by_i = \max \limits_{j \in \mN_e(i)} MLP \big( \bx_i, \bR_i^T(\bx_j - \bx_i) \big)
\label{eq:local_edgeconv_with_normals}
\end{equation}
In this manner, the network also considers differences of normal coordinates in a neighborhood around each point transformed in a local coordinate frame. 
We note that we also experimented with processing points together with normals as input to  the original EdgeConv  as first layer (instead of LocalEdgeConv). However, the gain was smaller compared to using LocalEdgeConv with normals as input features.
  
  \begin{figure*}[t!]
     \centering
     \includegraphics[width=1\textwidth]{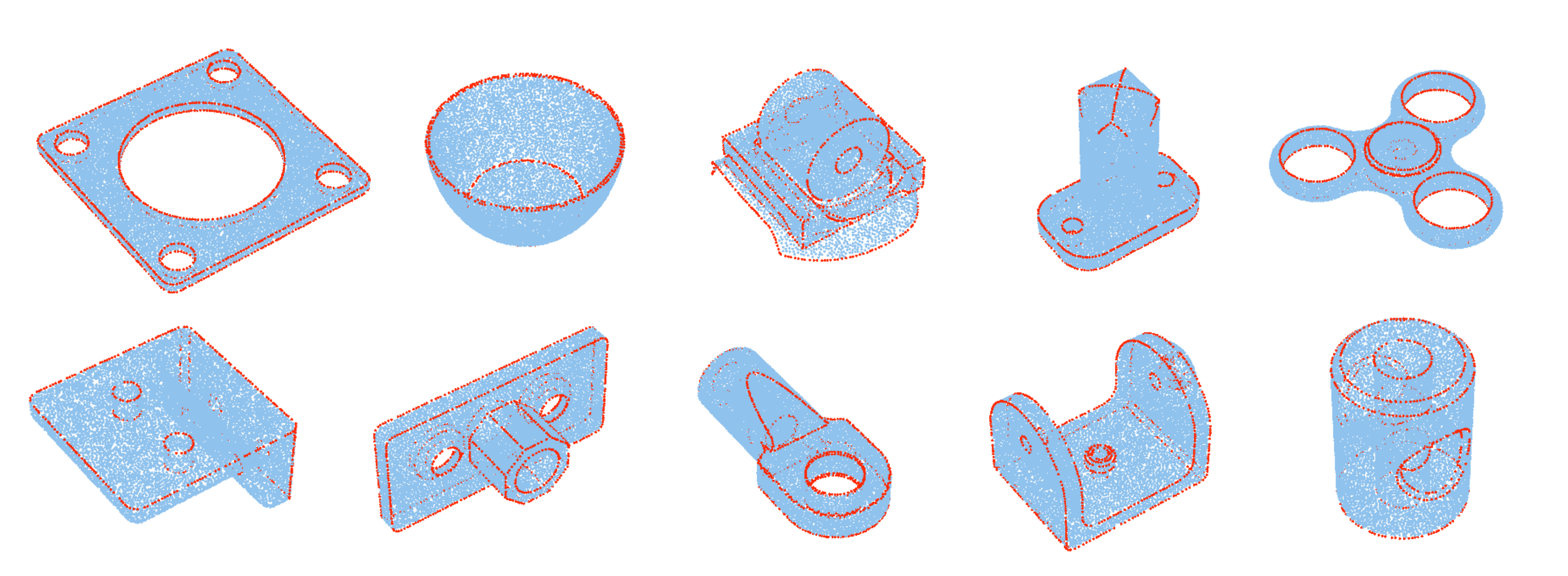}
\vspace{-5mm}     
     \caption{Marked (with red) boundaries on ABC point clouds for training.}
     \label{fig:abc_training_data}
\vspace{-5mm}     
 \end{figure*}
 
\paragraph*{Architecture.} After using a LocalEdgeConv layer (with or without normals as additional features), our architecture stacks two EdgeConv layers (Figure \ref{fig:architecture}) that sequentially process the representations extracted based on our local coordinate frames. At each EdgeConv layer, each point  is connected  its $K$ nearest neighbors dynamically updated  from the input feature space of the layer, as done in DGCNN \cite{Wang19}. The  point-wise representations extracted from the LocalEdgeConv and the two EdgeConv layers are concatenated, then processed through a max pooling layer, which produces a global shape descriptor.
The global descriptor is tiled and horizontally concatenated with the point-wise representations (Figure \ref{fig:architecture}),
so that the resulting point representations encode both local and global shape information. These are passed into a MLP, followed by a sigmoid transformation that outputs a boundary probability $b_i$ for each point $i$.

\subsection{Datasets}
\label{sec:datasets}  
To train our network, we make use of datasets that provide shape segmentations. We made use of two datasets for training and evaluation: a geometric segmentation dataset
and a semantic segmentation one, described below. We train and test our architecture on each dataset separately. 

\begin{figure*}[t!]
     \centering
     \includegraphics[width=1\textwidth]{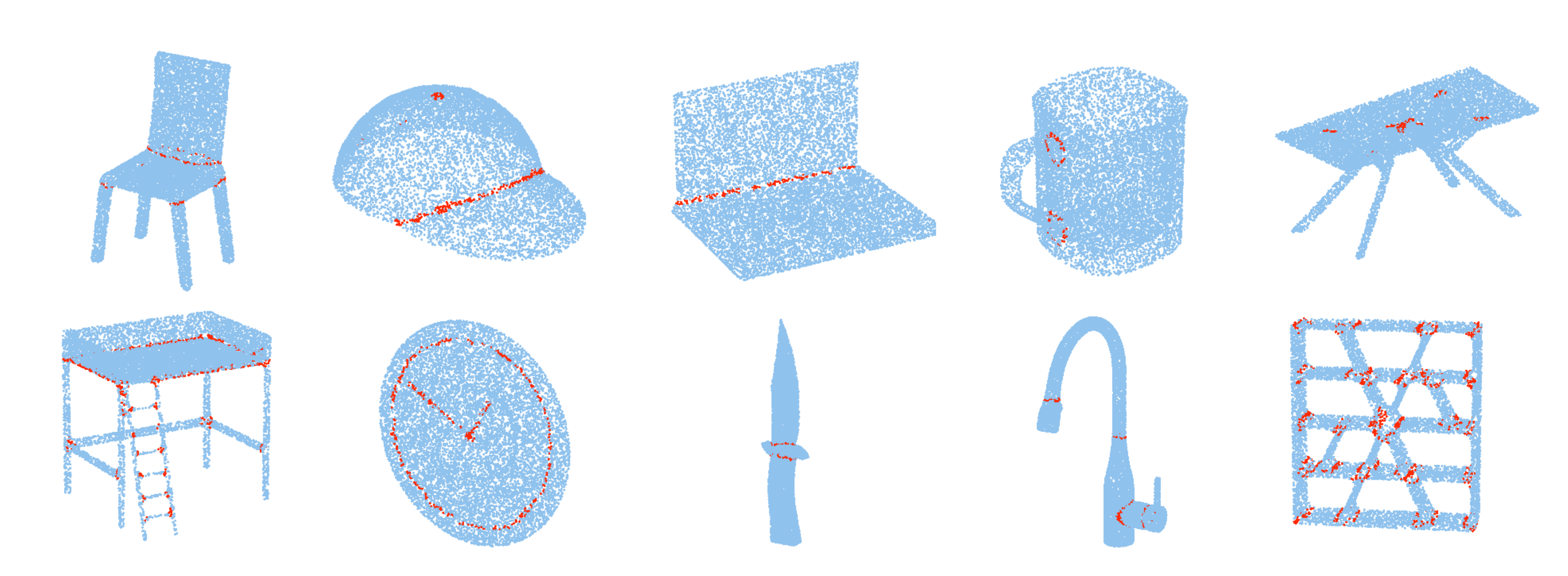}
\vspace{-5mm}        
     \caption{Marked boundaries on PartNet point clouds for training.}
     \label{fig:partnet_training_data}
\vspace{-5mm}        
 \end{figure*}

\paragraph*{Geometric segmentation dataset.}  The ABC dataset \cite{Koch2019} recently introduced a large repository of 3D\ geometric models, each 
defined by parametric surfaces and  ground truth information on their decomposition into individual patches. This dataset is a good source to learn segmentation boundaries between geometric primitives and patches. Another advantage of this dataset is that the patch boundaries are provided in parametric curve format, which allows us to extract very accurate  boundaries for training and evaluation. 

Since our goal is to detect boundaries for input point clouds, we first convert the geometric models into point-sampled surfaces for training. Specifically, we first  sample the surface of 3D models with $10K$  points based on Poisson-Disk sampling \cite{Ebeida11}, to create an initial point cloud.
Since it is rather unlikely to sample points lying exactly on boundary curves with this sampling procedure, we perform a second pass where we randomly sample another $10K$ points along boundary curves, specifically based on their underlying parametric representation. Concatenating the surface point samples of the first pass with the boundary point samples of the second pass tends to create higher point cloud density near the boundary regions.
To avoid this higher density bias during training, we perform a third pass where for each boundary point, we remove any surface samples within a distance equal to $\epsilon$, which is computed by measuring the distance of each point sample of the first pass to its nearest neighbor, then setting it to the maximum  distance over the point cloud.
Finally, we observed that some ABC shapes sometimes contain adjacent patches of same local geometry (e.g., two adjacent planes forming a flat boundary), where the boundary between them can be ignored. We filtered out such boundaries. 
All shapes are centered in the origin and scaled
so they lie inside the unit sphere.

The result of this procedure is the generation of a point cloud for each ABC shape with surface points carrying a binary label: boundary or non-boundary point.  Figure \ref{fig:abc_training_data} shows examples of such point clouds colored according to the binary label. We created a training set of $16,291$ labeled point clouds from ABC based on the above procedure. 

To monitor the training procedure, we also need a hold-out validation set. In addition, for evaluation, we need a test set. We gathered an additional set of $2,327$ shapes for hold-out validation and $4,655$ shapes for testing i.e., in total we had $23,273$  point clouds from ABC, and a $70\%$-$10\%$-$20\%$ proportion for training, validation and testing respectively. It is also important to note that the hold-out validation and testing point clouds are generated with Poisson point sampling from the original surfaces without adding boundary points (i.e., without the second and third pass used in training shapes). In this manner, we avoid biasing our testing procedure with point samples that are exactly at the geometric boundaries, and which may exhibit particular regular  patterns due to their sampling from the underlying parametric representation of boundary curves.  As discussed in our results section, the goal of our evaluation metrics is to detect boundaries up to a certain distance tolerance i.e., find points whose distance to the ground-truth parametric curve boundaries is up to distance equal to the maximum point sampling distance $\epsilon$. Finally, to simulate noisy point clouds for validation and testing, we add isotropic Gaussian noise to point coordinates with $\mu=0$ and $\sigma=0.005$. Normals are also perturbed from their original direction, by an angle sampled from a normal distribution trimmed within an interval $[-3, 3]$ degrees.

\paragraph*{Semantic segmentation dataset.} To learn boundaries for semantic segmentation, we use the recent PartNet dataset \cite{Mo_2019_CVPR}. The dataset provides  hierarchical segmentations of $26,671$ shapes into labeled parts in $24$ categories. The shapes are provided in the form of polygon meshes split into parts according to their label. We use the segmentations from the last hierarchy level in each category (i.e., the ``fine-grained'' segmentations). To generate the training point clouds with boundaries, we follow the following procedure.
First, we sample $10K$ points based on Poisson-Disk sampling. Partnet does not provide 
 boundary curves. Furthermore, neighboring parts in PartNet meshes are topologically disconnected from each other in their mesh representation,
often inter-penetrate each other, or even do not touch each other. We instead mark as boundaries all points of triangles that have neighboring points labeled with a different part label within a radius equal to $\epsilon$ set to the largest distance between all-pairs of nearest neighboring point samples. We add the same noise profile in the validation and test shapes, as in our ABC dataset.  
Figure \ref{fig:partnet_training_data} shows examples of the resulting point clouds, colored according to the binary label. The boundaries are more fuzzy and spread compared to the ABC\ dataset, yet are still clearly indicating zones separating semantic parts. 
 PartNet provides training, hold-out validation, and test splits, thus we follow the same splits in our case.

\subsection{Training procedure}
\label{sec:training}  
\label{sec:loss}

To train our architecture, we use the marked boundary and non-boundary points from either of the above training datasets as supervisory signal. We treat the problem as binary classification, and we use binary cross-entropy
as our loss function. However, since the number of boundary points is extremely small compared to the number of non-boundary ones (i.e., they represent less than $1\%$ of the total points on average), we use a weighted cross entropy loss that emphasizes the error on boundary points more: 
\begin{equation}
L=\sum\limits_{s \in \mD} \sum\limits_{i=1}^{N_s} 
w_{b} \cdot \hat t_i \cdot \log(b_i) +
 (1 - \hat t_i) \cdot  \log(1-b_i) 
\end{equation}
where $\hat t_i=1$ for marked boundary points and $\hat t_i=0$  for non-boundary points, and $w_b$ weights the cross-entropy terms for boundary points. Specifically, we set the weight according to the ratio of the number of non-boundary points and the number of boundary points:
$w_{b} = 
(\sum_i{[\hat t_i == 0]})/
(\sum_i{[\hat t_i == 1]})$.
In this manner, we penalize more misclassifications of boundary points. During training, as a form of data augmentation, and  to also increase robustness,
we add random noise in the points and normals (same noise distributions used in the validation and test sets of our datasets).

\paragraph*{Implementation details.}
Training  is  done  through  the Adam optimizer \cite{Kingma14} with learning rate $0.001$, beta coefficients set to $(0.9,0.999)$, batch normalization with momentum set to $0.5$ and batch normalization decay set to $0.5$ every $10$ epochs. The batch size is set to $8$ point clouds. Our  implementation is in Tensorflow and is publicly available at: \url{https://github.com/marios2019/learning_part_boundaries}.

\section{Evaluation}
\label{sec:eval}

\begin{figure*}[tbp]
     \centering
     \includegraphics[width=1\textwidth]{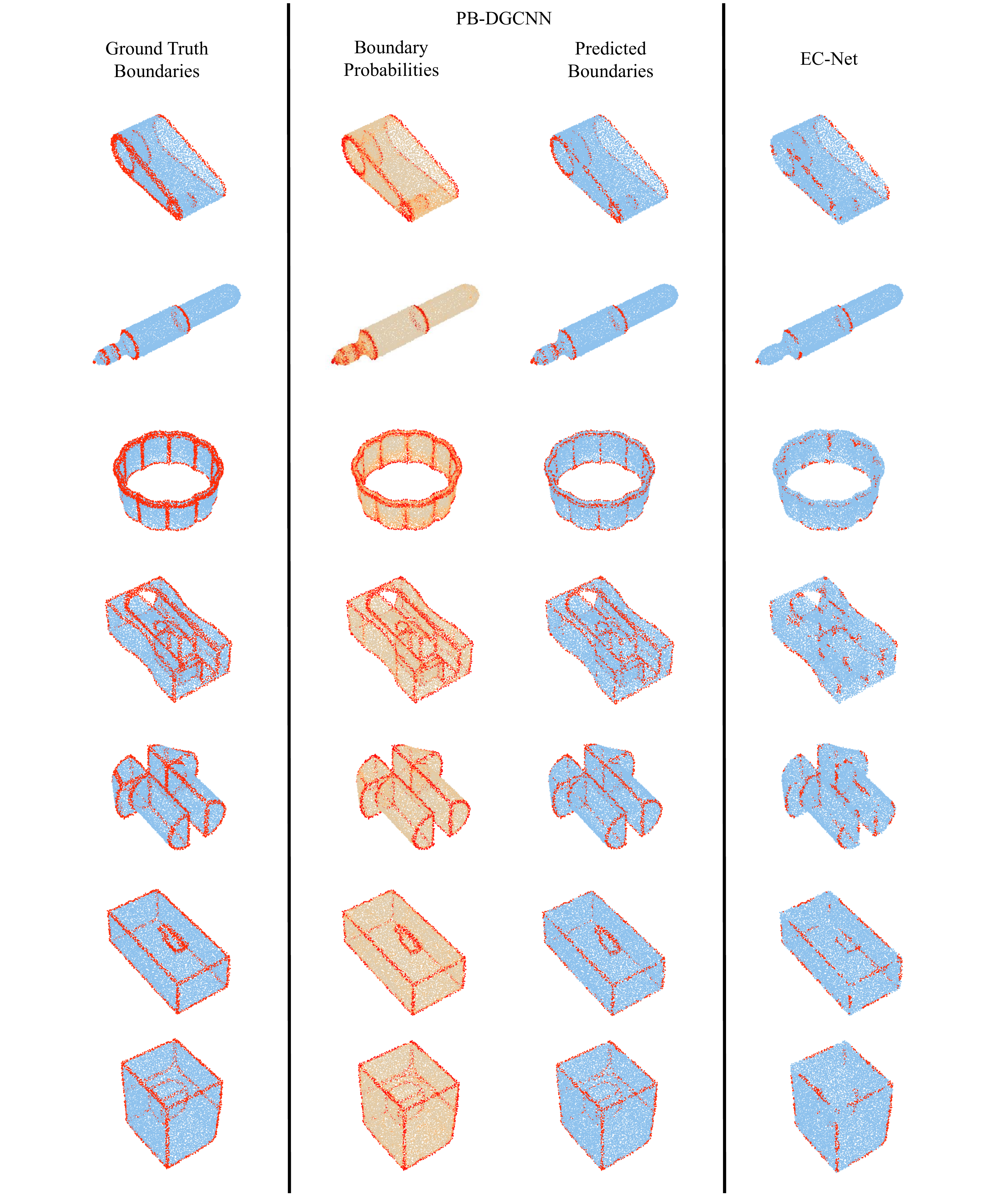}
     \caption{Visual comparison of the boundaries detected by our method PB-DGCNN, and EC-Net on some example ABC point clouds. The first column on the left shows the ground truth boundaries. The second column shows boundary probabilities produced by PB-DGCNN, and the third column shows boundaries predicted by PB-DGCNN after thresholding. The last column shows the boundaries predicted by EC-Net.}
     \label{fig:abc_gallery}
 \end{figure*}

\begin{table*}[tbp]
    \begin{center}
    \begin{adjustbox}{max width=\textwidth}
        \begin{tabular}{*{8}{c}}
        \toprule
        Model & \multicolumn{2}{c}{Input Features} & \multicolumn{5}{c}{Metrics}  \\
        \cmidrule(lr){1-1}
        \cmidrule(lr){2-3}
        \cmidrule(lr){4-8}
        & position & normal & CD & bIoU & F$_1$ & P & R   \\
        \midrule
        \multirow{ 2}{*}{EC-Net} & \checkmark & & 4.9 & 52.7 & 64.6 & 88.5 & 50.9 \\
                                 & \checkmark & \checkmark & 7.5 & 56.9 & 67.2 & 85.7 & 55.3 \\
        \cmidrule(lr){1-8}
        \multirow{ 2}{*}{PB-DGCNN w/ EdgeConv} & \checkmark & & 3.0 & 81.6 & 85.2 & 89.8 & 81.1 \\
                                                 & \checkmark & \checkmark & 2.1 & 89.8 & 90.3 & 90.9 & 89.7 \\
        \cmidrule(lr){1-8}
        \multirow{2}{*}{PB-DGCNN w/ LocalEdgeConv-curv} & \checkmark & & 2.6 & 85.2 & 88.0 & 91.3 & 85.8 \\
                                                     & \checkmark & \checkmark & 2.0 & 89.2 & 90.5 & \textbf{91.8} & 89.1 \\
        \cmidrule(lr){1-8}
        \multirow{2}{*}{PB-DGCNN w/ LocalEdgeConv} 
                                                     & \checkmark & & 2.4 & 90.0 & 89.7 & 89.2 & 90.1 \\
                                                     & \checkmark & \checkmark & \textbf{1.9} & \textbf{92.0} & \textbf{91.9} & \textbf{91.8} & \textbf{92.1} \\
        \bottomrule
        \end{tabular}
    \end{adjustbox}
    \end{center}
    \caption{Boundary classification results on the ABC dataset (CD: Chamfer Distance - \%, bIoU: Boundary IoU - \%, F$_1$: F$_1$ score - \%, P: Precision - \%, R: Recall - \%)}
    \label{table:abc_evaluation}
\vspace{-5mm}        
\end{table*}

We now discuss experimental evaluation of our method. First, we introduce evaluation metrics for part boundary detection, and present results on the ABC dataset for geometric boundary detection. Then we present an application of our method to semantic segmentation, and present evaluation on the PartNet dataset.

\subsection{Evaluation metrics}

Our evaluation metrics are inspired by the literature on line drawing and segmentation for 3D meshes. Cole et al. \cite{Cole:2008} introduced metrics that evaluate similarity of human-annotated line drawings with computer-generated ones based on precision and recall. Liu et al. \cite{Liu_2020_CVPR} extended these metrics to include Intersection over Union (IoU). Our part boundaries can be thought of as point-sampled lines in 3D, thus we also use precision, recall, and IoU inspired by these works.
Chen et al \cite{Chen:2009:ABF} introduced various metrics for evaluating segmentation for 3D\ meshes. In the case of boundaries, they propose cut discrepancy that measures distances of annotated and predicted boundaries on the surface. 
Following the above  works,
we introduce the following metrics for evaluation boundaries:

\paragraph*{Precision} is defined as the fraction of predicted boundary points in a point cloud that are ``near'' any annotated boundary. The proximity is computed by measuring Euclidean distance of points to boundary curves in ABC dataset, or boundary point samples in PartNet. The definition of ``near'' requires a distance threshold indicating tolerance to small errors. We define this tolerance as the maximum point sampling distance $\epsilon$ (largest distance between all-pairs of nearest neighboring point samples per point cloud). We also examine performance under varying levels of tolerance (multiples of $\epsilon$). 
\vspace{-3mm}

\paragraph*{Recall} is defined as the fraction of annotated boundary points that are ``near'' any predicted boundary point. We follow the same definition of nearness as above. In the ABC dataset, we densely sample the parametric boundary curves to evaluate recall. \vspace{-3mm}

\paragraph*{F1-score} is the harmonic mean of precision and recall, often used to combine them both in one metric. \vspace{-3mm}

\paragraph*{Boundary IoU (bIoU)} is the  Intersection over Union  that measures ``overlap'' between annotated boundaries and predicted ones. A boundary and predicted point ``overlap'' if they are near to each other, based on the same definition of nearness as above. \vspace{-3mm}
 
\paragraph*{Chamfer distance (CD)} measures Euclidean distance from annotated boundary samples to nearest predicted boundary points, and vice versa (i.e., we use the symmetric Chamfer distance).\\

\begin{figure*}[t!]
     \centering
     \includegraphics[width=1\textwidth]{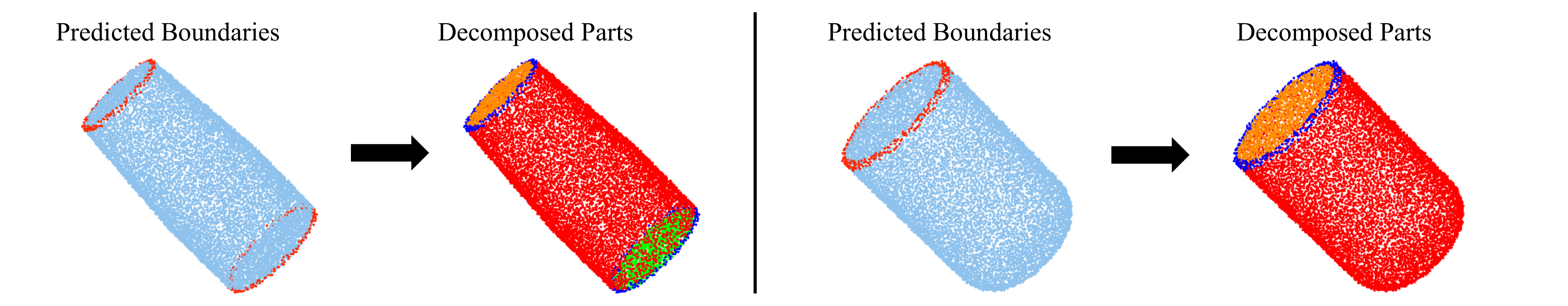}
\vspace{-5mm}     
     \caption{Examples of watershed (flood-filling) segmentation. In these cases well-defined predicted boundaries between geometric parts, enable their decomposition to individual segments through simple BFS-based flood-filling.}
     \label{fig:watershed_segmentation}
\vspace{-3mm}     
 \end{figure*}

\noindent It is important to note that in order to evaluate the above metrics, the probabilistic boundaries must be binarized first. In the ABC\ dataset, we use thresholding (i.e., a point becomes boundary if its probability is above a threshold). To select the threshold, we perform dense grid search in our hold-out validation dataset, and select the value that minimizes the Chamfer Distance. In the PartNet dataset, the probabilistic boundaries are used in a pairwise term in graph cuts - the points crossed by the cut are marked as boundaries. Finally, we note that the metrics are computed for each test point cloud shape, then averaged over the test shapes.

\begin{figure}[t!]
     \centering
     \includegraphics[width=0.45\textwidth]{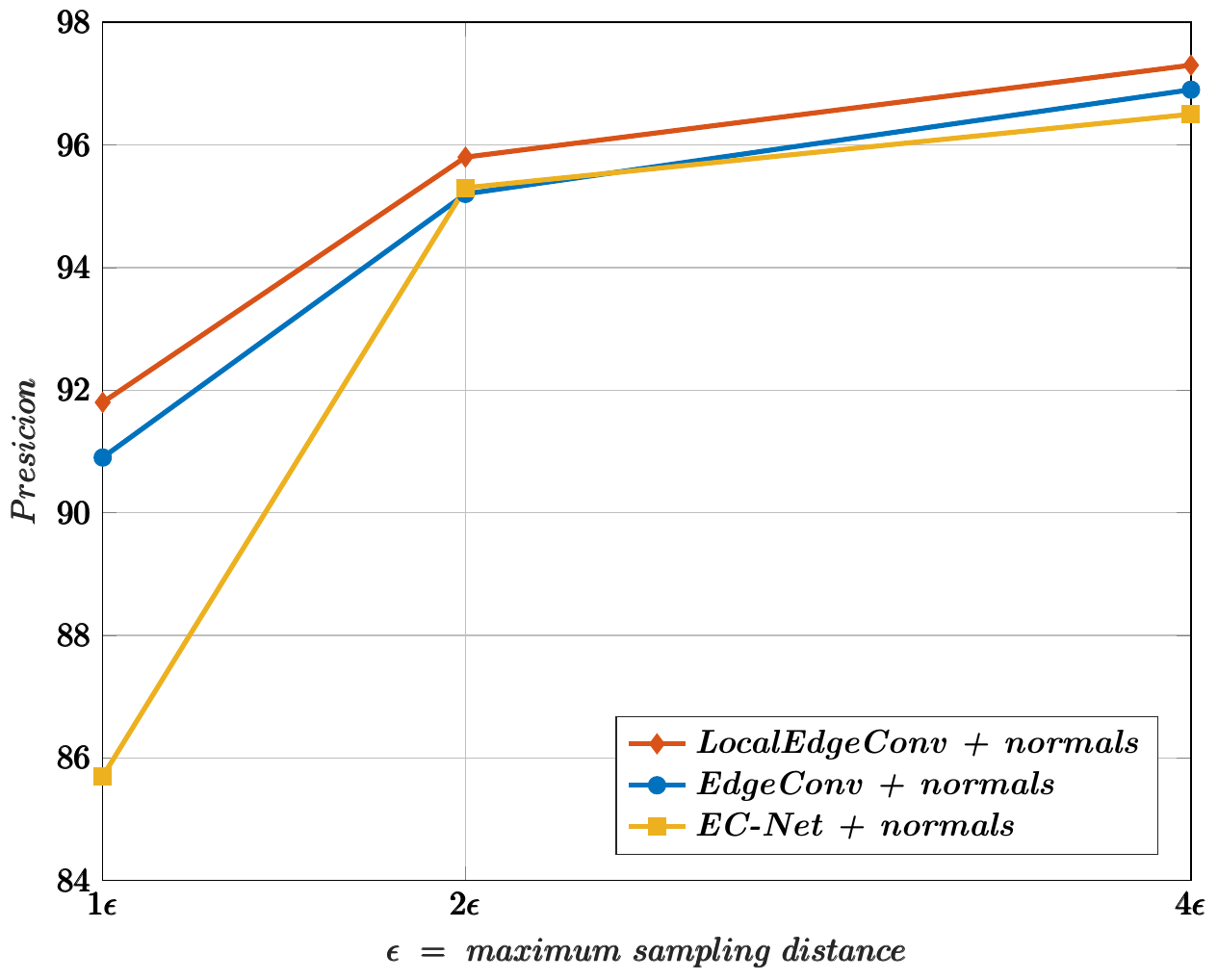}
     \includegraphics[width=0.45\textwidth]{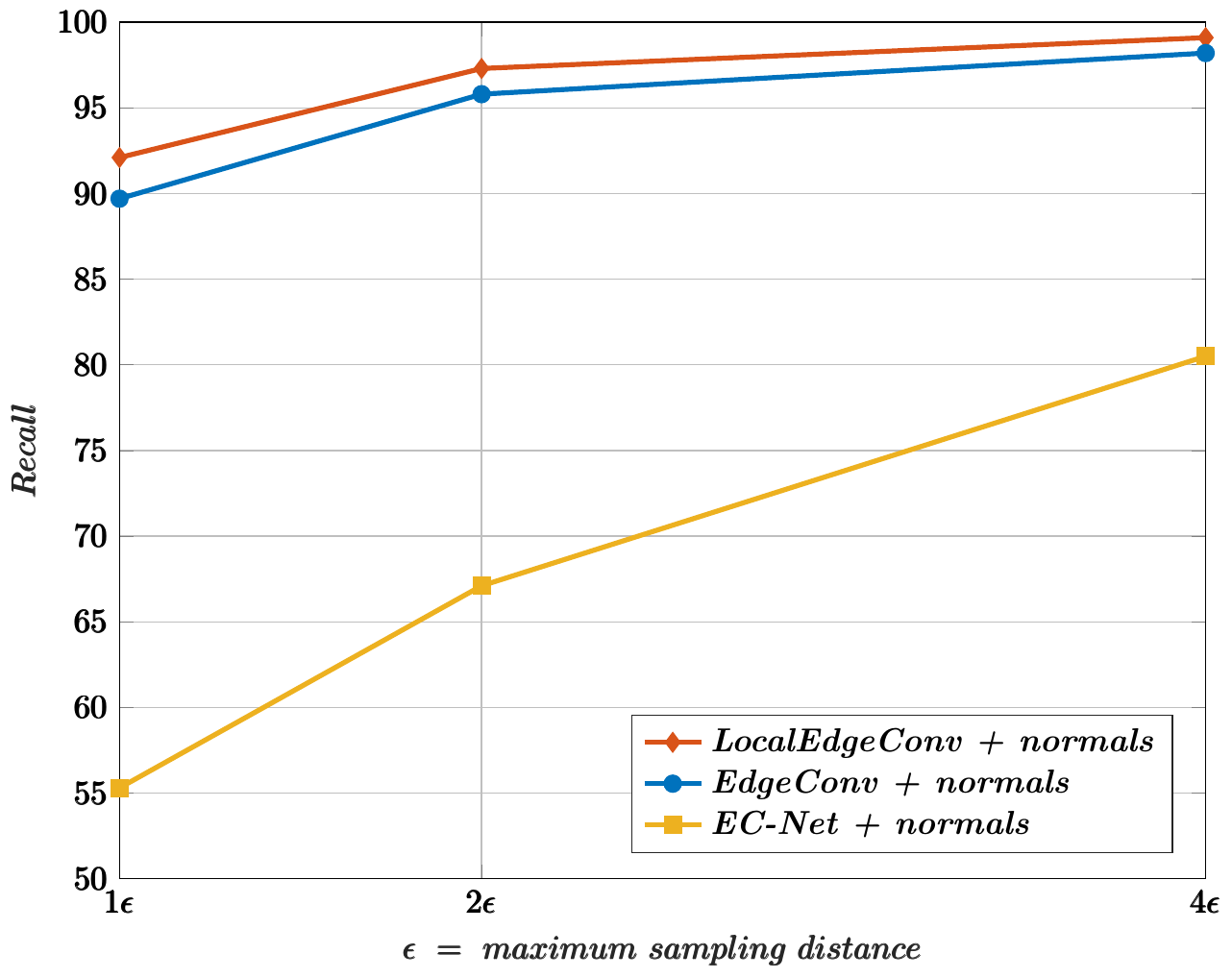}
\vspace{-2mm}     
     \caption{Boundary detection evaluation wrt. precision and recall for different tolerance levels.}
     \label{fig:plot_metrics}
\vspace{-5mm}     
 \end{figure}

\begin{figure*}[t!]
     \centering
     \includegraphics[width=1\textwidth]{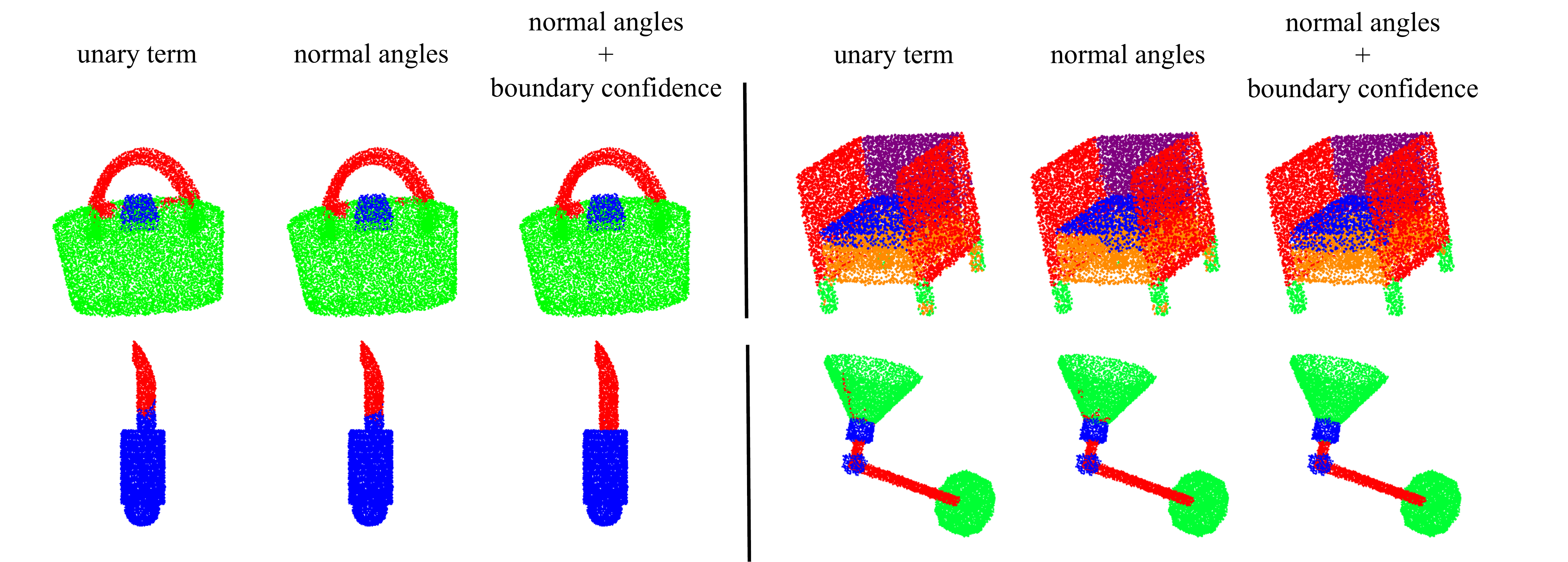}
\vspace{-5mm}     
     \caption{Visual comparison of semantic segmentation for example PartNet point clouds, using DGCNN alone (unary), a graph-cut formulation with normal angles in the pairwise term, and a graph-cut formulation with a combination of normal angles and boundary probabilities produced by PB-DGCNN in the pairwise term. Boundary confidence can help to diminish small semantic segments, which are falsely predicted from the semantic segmentation model (unary). These are the cases of Bag (top row, left) and Lamp (bottom row, right), where wrong annotated parts are present even after applying the graph cuts method with normal angles as a pairwise term.  Moreover, it can further smooth out semantic parts as it is been illustrated in the case of Chair (top row, right) and Knife (bottom row, left).}
     \label{fig:semantic_segmentation}
\vspace{-2mm}  
\end{figure*}

\subsection{Geometric part boundary detection}

We now discuss evaluation for detection of part boundaries between geometric primitives on ABC \cite{Koch2019} based on the dataset described in Section \ref{sec:datasets}. The primitives in ABC include plane, cone, cylinder, sphere, torus, surface of revolution or extrusion or NURBS patch. We compare our method with the edge detection network called EC-Net from \cite{yu2018ec}. The method was introduced for detecting edges on point clouds for 3D reconstruction. It upsamples the original point cloud, while we also producing a value per point corresponding to its distance to the nearest edge. By thresholding the value, the method detects edges. We adapted their method for our task. We trained their method on our dataset, tuned their hyper-parameters
(weights of losses) in our hold-out validation set, tuned the threshold for edge detection using hold-out validation to optimize Chamfer distance, and used the same augmentation as in our method. Since their method is based on sampling individual patches from the point cloud, we experimentally verified  that the sampled patches fully cover the ABC shapes by setting their number to $50$. 

Table \ref{table:abc_evaluation} reports our five evaluation metrics for EC-Net and our method. We evaluated two version of EC-Net: one with points only as input features (EC-Net w/o normals), and another with points and normals as input features
(EC-Net w/ normals).
As indicated by all metrics, our method produces boundaries that are much closer to the annotated ones compared to EC-Net. For example, the EC-Net without normals has $2.04$ times higher error than our method without normals (see PB-DGCNN w/ LocalEdgeConv w/o normals) in terms of Chamfer Distance, and
$2.58$ times higher error than our method with normals (see PB-DGCNN w/ LocalEdgeConv w/ normals). The EC-Net with normals seems to have even higher error in Chamfer Distance, yet better Recall and IoU profile than EC-net without normals. It seems that the EC-Net with normals makes better predictions near ground-truth boundaries, but also produces additional boundaries away from ground-truth ones, which results in higher Chamfer Distance. In any case, our PB-DGCNN with LocalEdgeConv offers much better performance compared to both versions of EC-Net according to all our evaluation metrics.

\paragraph*{Ablation study.}
In Table \ref{table:abc_evaluation}, we also report the performance of our method under the following variants: 
(a) ``PB-DGCNN w/ EdgeConv'' where we use the original EdgeConv layer of DGCNN \cite{Wang19} as first layer instead of LocalEdgeConv. We include the performance of this variant with and without using normals as input features (b) ``PB-DGCNN w/ LocalEdgeConv-curv'' where we use the principal curvature directions as tangent vectors $\bu_i$ and $\bv_i$ to define the local coordinate frame $\bR_i$ per point (we note that since the curvature directions are defined up to a sign, the max operator in Eq. \ref{eq:local_edgeconv_with_normals} is applied to MLPs that also include coordinate transformations based on the opposite principal directions). Curvature is estimated based on the method proposed in \cite{Kalogerakis:2007:robustCurvature}.
Finally, we include the performance of of our method ``PB-DGCNN w/LocalEdgeConv'' with and without normals as input features. Based on the numerical results, we observe that ``PB-DGCNN w/LocalEdgeConv w/ normals'' has the best performance on average. Its achieved Chamfer Distance is lowest compared to all variants, and the bIoU and F1 score are the highest. Using principal directions did not seem to help the performance of LocalEdgeConv. The LocalEdgeConv w/ normals  has consistently better  performance compared to using EdgeConv w/ normals according to all metrics. Similarly LocalEdgeConv w/o normals is better than using EdgeConv w/o normals on average. The precision of EdgeConv w/o normals is a bit higher than LocalEdgeConv w/o normals, yet note that its recall is much lower.

Figure \ref{fig:plot_metrics} shows precision and recall for LocalEdgeConv, EdgeConv and EC-Net (here, we use points and normals as input to all methods). Increasing the tolerance results in increasing the precision for all methods, since more predicted boundaries are classified as correct by increasing the boundary distance tolerance threshold, as expected. Similarly, recall is also increased.
Most importantly, our method based on LocalEdgeConv has better behavior than the rest, since it demonstrates both higher precision and recall for all tolerance levels we examined.

Figure \ref{fig:abc_gallery} provides a visual demonstration for some example point clouds from ABC. We find that the EC-Net boundaries are highly noisy and inconsistent, while ours tend to agree with the ground-truth more.

\begin{figure*}[t!]
     \centering
     \includegraphics[width=1\textwidth]{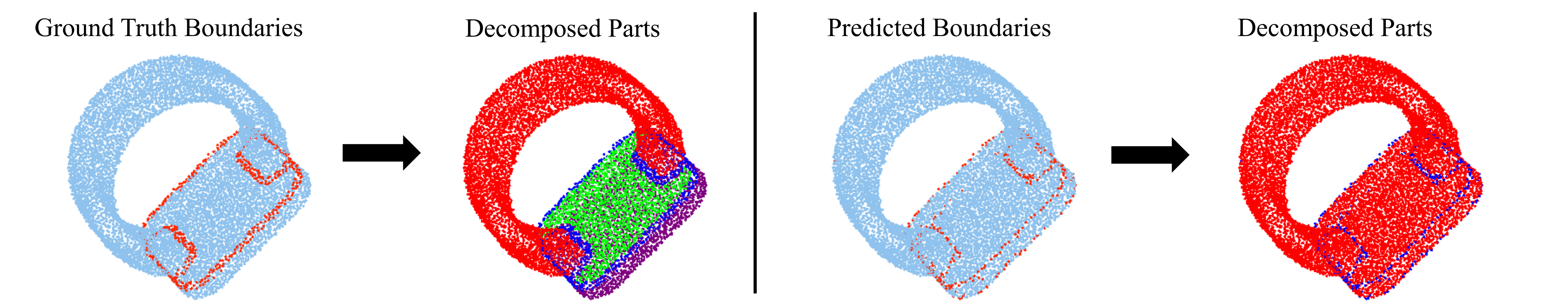}
\vspace{-5mm}        
     \caption{The figure on the left depicts part decomposition of the point cloud from ground truth boundaries, with BFS flood-filling. It successfully segments the point cloud into three parts. This is not the case on the right figure, where the predicted boundaries fail to enclose points into separate segments, which results to only one part after the watershed segmentation procedure.}
     \label{fig:limitation}
\vspace{-5mm}        
 \end{figure*}

\paragraph*{Application to geometric decomposition.} We found that our boundaries can be used for segmentation of several ABC shapes using a simple flood-filling, watershed segmentation approach
(Figure \ref{fig:teaser} and \ref{fig:watershed_segmentation}). We first construct a K-NN graph ($K=4$) over the point cloud, then we perform a BFS starting from a random seed point and stopping at predicted boundary points. All visited points result in a segment. Then we start the same procedure by using a random seed point from the rest of the non-visited points. We note, however, that this simple flood-filling approach can fail in cases where small gaps exist in boundaries (Figure \ref{fig:limitation}).


  \begin{table*}[t!]
    \begin{center}
    \begin{adjustbox}{max width=\textwidth}
        \begin{tabular}{*{26}{c}}
        \toprule
        \textbf{Category } & Bag & Bed & Bott & Bowl & Chai & Cloc & Dish & Disp & Door & Ear & Fauc & Hat & Key & Knif & Lamp & Lap & Micr & Mug & Frid & Scis & Stor & Tabl & Tras & Vase & \textbf{Avg} \\
        \midrule
        \textbf{Shape IoU} \\
        \cmidrule{1-1}
        Unary only & 75.9 & 25.7 & 59.1 & 75.5 & 50.8 & 43.9 & 53.9 & 84.0 & 44.0 & 52.9 & 55.8 & 64.3 & \textbf{62.4} & 43.3 & 43.5 & 95.9 &     59.9 & 88.4 & 51.6 & 76.8 &        52.1 & 52.9 & 52.4 & 80.8 &     60.2 \\        
                GC normal diff & 76.1 & 25.9 & 60.6 & 81.3 &    55.0 & \textbf{44.1} & 54.1 & 85.2 & 45.4 & \textbf{53.0} & 58.0 & 65.2 &    \textbf{62.4} & 45.6 & 47.2 & \textbf{96.0} &       60.7 & 89.5 & \textbf{52.9} & 76.9 &    55.2 & 55.7 & 54.0 & 82.5 &        61.8 \\
        GC PB-DGCNN & 76.1 & 26.1 & 61.4 & 83.2 &       54.6 & 43.7 & 54.0 & \textbf{86.1} & 48.9 & 52.9 & 57.5 & 70.3 &   \textbf{62.4} & 47.4 & 48.5 & 94.6 &    \textbf{60.9} & \textbf{90.4} & 51.6 & 77.2 & 54.1 & 56.3 & \textbf{55.0} & 83.5 & 62.4 \\
        GC both & \textbf{76.2} & \textbf{26.2} & \textbf{61.6} & \textbf{84.6} &       \textbf{56.3} & 43.7 & \textbf{54.4} & 85.9 & \textbf{49.3} & 52.9 & \textbf{58.4} & \textbf{72.2} & \textbf{62.4} & \textbf{48.0} & \textbf{49.9} & 94.6 & 60.8 & 88.7 &  52.6 & \textbf{78.0} & \textbf{55.3} & \textbf{57.0} & 54.7 & \textbf{83.8} & \textbf{62.8} \\
        \midrule
        \textbf{Part IoU} \\
        \cmidrule{1-1}
        Unary only & 49.9 & 26.8 & 39.9 & 64 & 40.6 & \textbf{24.6} & 46.2 & 84.3 & 32.2 & \textbf{42.4} & 46.1 & 62.7 & \textbf{61.1} & 39.5 & 24.1 & 95.6 & \textbf{54.4} & 81.5 & 37.7 & 76.5 & 43.1 & 33.4 & \textbf{45.5} & 55.9 & 50.3 \\
        GC normal diff & 50.0 & \textbf{27.0} & 39.9 & 64.2 &   41.5 & \textbf{24.6} & 46.5 & \textbf{84.6} & 32.7 & \textbf{42.4} & 47.1 & 63.0 &   \textbf{61.1} & 38.5 & 24.3 & \textbf{95.7} &        52.3 & 80.4 & \textbf{38.1} & 76.6 &    \textbf{43.3} & \textbf{33.6} & 42.8 & \textbf{56.8} &        50.3 \\
        GC PB-DGCNN & \textbf{50.2} & \textbf{27.0} & 44.3 & 66.7 &     41.2 & 24.0 & 46.7 & \textbf{84.6} &        32.6 & \textbf{42.4} & 46.7 & 63.6 &    \textbf{61.1} & \textbf{39.7} & \textbf{24.4} & 93.8 &        53.9 & \textbf{82.7} & 37.7 & 76.8 &       43.2 & 33.4 &  43.8 & 56.6 & 50.7 \\
        GC both & \textbf{50.2} & \textbf{27.0} & \textbf{45.0} & \textbf{69.4} &       \textbf{41.6} & 24.0 & \textbf{47.0} & 84.4 & \textbf{33.1} & \textbf{42.4} & \textbf{47.3} & \textbf{65.7} & \textbf{61.1} & 38.4 & \textbf{24.4} & 93.9 & 52.2 & 80.2 & \textbf{38.1} & \textbf{77.6} & 43.2 & 33.4 & 42.0 & 56.7 & \textbf{50.8} \\
        \bottomrule
        \end{tabular}
    \end{adjustbox}
    \end{center}
    \caption{Point labeling evaluation of fine-grained semantic segmentation on the PartNet dataset (Part IoU, Shape IoU - \%).
``Unary alone'' represents using the per point part probabilities produced by DGCNN,  ``GC normal diff''
represents graph cuts using normal angles as pairwise term, ``GC PB-DGCNN'' represents graph cuts using our predicted boundary confidences as pairwise term, and ``GC both`` represents graph cuts using the weighted combination of pairwise terms based on both normal angles and PB-DGCNN.}
    \label{table:semantic_segmentation_iou_results}
\end{table*}

\subsection{Semantic shape segmentation}
\label{sec:segmentation}

We now discuss evaluation on semantic shape segmentation based on the PartNet dataset. Here, we train our network on the PartNet training split for each of its categories, as described in Section \ref{sec:datasets}. To take advantage of semantic part labels, here we first use a network that predicts a probability for each part per point. Specifically, we use the DGCNN network for this task \cite{Wang19}, operating on  10K number of point samples per shape. We then incorporate a graph cuts formulation, where the above per-point part probability is used as a unary term, and the output boundary probabilities from our method (``PB-DGCNN w/ LocalEdgeConv w/normals'') are used as a pairwise term:
\begin{equation}
    E(\bc) = \sum_{i \in \mP} \psi(c_i) + \sum_{i \in \mathcal{P}}
\sum_{j \in \mN(i)} \phi(c_i, c_j)
\label{eq:graph_cuts}
\end{equation}
where $\bc=\{c_i\}$ are the label assignments we wish to compute by minimizing the above energy, $\mP$ is the set of points in a test point cloud, and $\mN(i)$ is the neighborhood of each point $i$ formed by its $K=4$ nearest Euclidean neighbors. The unary term is expressed as follows $\psi(c_i)=-log P(c_i)$, where $P(c_i)$ is the probability distribution over part labels associated with the point $i$ produced by the DGCNN point labeling network. The pairwise term uses the maximum of our PB-DGCNN boundary probabilities for the two points:  $\phi(c_i, c_j)=- \lambda \cdot log (\max(b_i, b_j))$ for $c_i \ne c_j$, and $0$ otherwise.  The weighting parameter $\lambda$ is adjusted through grid search in the hold-out validation set  per shape category.
To avoid infinite costs, we add a small $\epsilon=10^{-3}$ to the above log expressions.

We also experimented with another pairwise term variant as baseline that considers angles between between point normals:\\
 $\phi'(c_i, c_j)=-\lambda' \cdot log ( min(\omega_{i,j}/ 90^o,1) )$,
 for $c_i \ne c_j$,  where $\omega_{i,j}$ is the angle between the point normals. The term results in zero cost for right angles between normals indicating a strong edge. The weighting parameter $\lambda'$ is adjusted through grid search in the hold-out validation set per shape category. We finally experimented with a combination of using both the above pairwise terms in Eq. \ref{eq:graph_cuts}: $\phi(c_i, c_j)  + \phi'(c_i, c_j)$.

We first report point labeling performance in Table \ref{table:semantic_segmentation_iou_results} based on the standard part IoU and shape IoU metrics in PartNet \cite{Mo_2019_CVPR}. We examine the performance of using DGCNN alone as unary term (``unary alone''), then using graph cuts based on the normal angle baseline described above (``GC normal diff''), graph cuts based on the predicted boundary probabilities of PB-DGCNN (``GC PB-DGCNN''), and finally graph cuts using the summation of pairwise terms from normal angles and PB-DGCNN (``GC both''). We observe small but noticeable average performance increases for both shape IoU and part IoU when using all variants of graph cuts. The best performance is achieved on average (see last row, last column) when combining both pairwise terms. 
Specifically, we observe an increase of average shape IoU by $2.6\%$ and part IoU $0.5\%$ compared to using the unary term alone. 
 We believe that using both pairwise terms offers the best performance because our boundary probabilities are more fuzzy in PartNet - we note that the training boundary data were also slightly fuzzy in PartNet (see Figure \ref{fig:partnet_training_data}) in the first place. Using normal angles further sharpens our probabilistic boundaries. 
Nevertheless, the metrics seem improved with the use of our probabilistic boundaries in the pairwise term alone. 
We note that graph cuts is executed in a deterministic manner (i.e., there is no variance in the above increases given a fixed unary term). The performance increase is not dramatic: this is not surprising, since refining boundaries
changes  relatively few point labels near boundaries. 

\begin{table*}[t!]
    \begin{center}
    \begin{adjustbox}{max width=\textwidth}
        \begin{tabular}{*{26}{c}}
        \toprule
        \textbf{Category} & Bag & Bed & Bott & Bowl & Chai & Cloc & Dish & Disp & Door & Ear & Fauc & Hat & Key & Knif & Lamp & Lap & Micr & Mug & Frid & Scis & Stor & Tabl & Tras & Vase & \textbf{Avg} \\
        \midrule
        \textbf{Chamfer Distance} \\
        \cmidrule{1-1}
        Unary alone & \textbf{6.3} & \textbf{3.1} &	\textbf{6.9} & 38.8 & 4.0 & \textbf{5.4} & 6.6 & 7.8 & 37.3 &	\textbf{9.9} & 6.0 &	4.1 & \textbf{2.3} &	9.4 &7.2 & \textbf{1.0} & \textbf{3.3} & 3.6 & \textbf{3.9} & 9.4 & 1.5 & 3.2 & \textbf{3.8} & \textbf{10.9} & 8.2 \\
        GC both & 6.4 & \textbf{3.1} &	7.1 & \textbf{37.0} & \textbf{3.7} & \textbf{5.4} & \textbf{6.2} & \textbf{7.7} & \textbf{36.4} &	\textbf{9.9} & \textbf{5.4} &	\textbf{3.1} & \textbf{2.3} &	\textbf{7.3} &\textbf{5.9} &	1.3 & 3.5 & \textbf{3.2} & \textbf{3.9} &	\textbf{8.7} & \textbf{1.3} &	\textbf{2.7} & 5.9 &	11.0 & \textbf{7.9} \\
        \midrule
        \textbf{Boundary IoU} \\
        \cmidrule{1-1}
        Unary alone & \textbf{61.5} & 72.3 & 48.0 & 56.0 & 67.0 & 59.0 & 64.0 & 73.5 & 54.4 &	\textbf{49.1} & 54.4 & 77.3 & \textbf{76.5} &	38.7 & 57.1 & 90.5 & \textbf{79.1} &	66.7 & 69.1 & 44.7 & 88.4 & 75.9 & \textbf{77.6} & 73.4 &	65.6 \\
        GC both & 60.9 & \textbf{73.0} & \textbf{48.4} & \textbf{60.7} & \textbf{70.2} & \textbf{63.9} & \textbf{66.0} & \textbf{74.9} & \textbf{57.9} & 49.0 & \textbf{63.2} & \textbf{82.1} & 76.1 & \textbf{47.0} & \textbf{67.0} & \textbf{90.9} & 76.4 &	\textbf{75.3} & \textbf{69.7} & \textbf{50.4} & \textbf{90.0} & \textbf{79.2} & 74.8 & \textbf{75.4} & \textbf{68.4} \\
        \midrule
         \textbf{Precision} \\
         \cmidrule{1-1}
         Unary alone & 74.3 & 70.5 & 52.3 &	58.7 & 65.3 & 57.8 & 67.1 & 72.1 & 56.9 & 56.6 &	51.8 & 78.4 & 81.8 & 34.8 &	56.2 & 91.2 & 81.4 & 75.0 &	76.4 & 46.2 & 88.6 & 78.0 & 78.4 & 72.8 & 67.6 \\
         GC both & \textbf{77.6} & \textbf{73.7} & \textbf{60.4} & \textbf{68.0} & \textbf{80.2} & \textbf{73.9} & \textbf{75.7} & \textbf{81.4} & \textbf{65.7} & \textbf{57.6} & \textbf{71.2} & \textbf{89.2} & \textbf{82.1} & \textbf{54.7} & \textbf{81.9} & \textbf{93.8} &	\textbf{81.7} & \textbf{87.3} & \textbf{80.3} & \textbf{54.8} &	\textbf{94.4} & \textbf{90.1} & \textbf{89.6} & \textbf{78.8} & \textbf{76.8} \\
         \midrule
       \textbf{Recall} \\
         \cmidrule{1-1}
         Unary alone & \textbf{57.9} & \textbf{79.5} & \textbf{48.9} & \textbf{62.8} & \textbf{74.7} & \textbf{68.0} & \textbf{68.0} & \textbf{79.1} & \textbf{56.1} & \textbf{47.8} & \textbf{61.8} & \textbf{78.9} & \textbf{75.9} & \textbf{51.6} & \textbf{72.5} & \textbf{91.0} & \textbf{78.0} & 65.1 &	\textbf{66.9} & 45.3 & \textbf{90.0} & \textbf{81.1} & \textbf{79.9} & \textbf{79.9} & \textbf{69.2} \\
         GC both & 55.9 & 76.8 & 44.8 &	62.1 & 67.1 & 63.4 &	63.2 & 72.7 & 55.9 & 46.8 &	60.2 & 78.0 & 75.3 & 46.2 &	63.9 & 90.0 & 73.3 & \textbf{70.6} & 65.4 & \textbf{48.9} & 87.5 & 76.2 & 68.9 & 76.4 & 66.2 \\
       \midrule
        \textbf{F$_{1}$-score} \\
        \cmidrule{1-1}
        Unary alone & \textbf{65.1} & 74.7 & 50.5 & 60.7 & 69.7 & 62.5 & 67.5 & 75.4 & 56.5 & \textbf{51.8} & 56.4 & 78.7 & \textbf{78.8} &	41.5 & 63.3 & 91.1 & \textbf{79.7} &	69.7 & 71.3 & 45.7 & 89.3 & 79.5 & \textbf{79.1} & 76.2 &	68.1 \\
        GC both & 65.0 & \textbf{75.2} & \textbf{51.}4 & \textbf{64.9} & \textbf{73.1} & \textbf{68.3} & \textbf{68.9} & \textbf{76.8} & \textbf{60.4} & 51.6 & \textbf{65.2} & \textbf{83.2} & 78.5 & \textbf{50.1} & \textbf{71.8} & \textbf{91.9} & 77.2 &	\textbf{78.0} & \textbf{72.1} & \textbf{51.7} & \textbf{90.8} & \textbf{82.5} & 77.9 & \textbf{77.6} & \textbf{71.0} \\
        \bottomrule
        \end{tabular}
    \end{adjustbox}
    \end{center}
    \caption{Evaluation of fine-grained semantic segmentation boundaries (Chamfer distance, Boundary IoU, Precision, Recall, F$_1$ score - \%) on the PartNet dataset. ``Unary alone'' represents using the per point part probabilities produced by DGCNN,  ``GC both'' represents a weighted combination of pairwise terms based on normal angles and PB-DGCNN.}
    \label{table:semantic_segmentation_results}
\vspace{-2mm}         
\end{table*}

Table \ref{table:semantic_segmentation_results}  reports our evaluation metrics in terms of boundary quality. We note that compared to the ABC dataset, the evaluation of boundaries here is less reliable. In contrast to ABC, where the ground-truth boundaries were parametric curves and were highly accurate, the ground-truth boundaries in PartNet are marked approximately using the heuristic search described in Section \ref{sec:datasets}. We report the performance of our best variant of graph cuts (using both terms), and also the unary term alone. We observe that graph cuts result in boundaries that are more consistent with ground-truth. In particular, we observe an improvement of $2.8\%$ for bIoU, and $2.9\%$ for F1-score on average. We note that although the recall is lower, the precision is significantly much higher. Figures \ref{fig:teaser} and \ref{fig:semantic_segmentation} show semantic segmentation results for a few examples from PartNet.



\begin{figure}[tbp]
     \centering
     \includegraphics[width=0.5\textwidth]{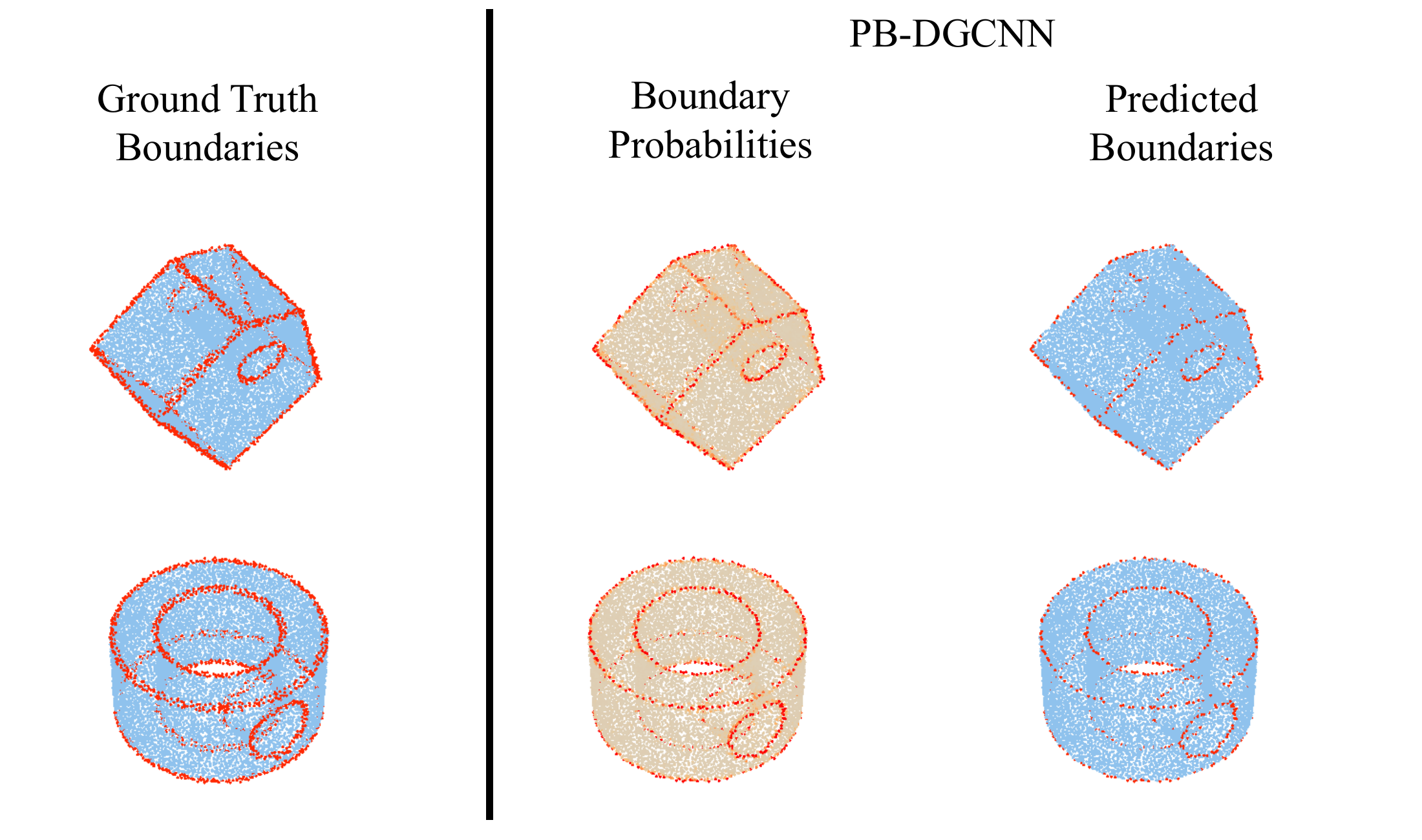}
     \caption{Predicted boundary confidences (middle column) is sometimes low resulting in sparsely labeled boundary points (right column).}
     \label{fig:part_boundaries_failure_cases}
     \vspace{-5mm}
 \end{figure}
 
\section{Limitations and Conclusion}

We presented a method for detecting probabilistic boundaries in point clouds based on a neural network. Our evaluation showed that our boundaries are closer to ground-truth in geometric decomposition tasks, and also improve the quality of cuts in semantic segmentation tasks. Our method also has limitations that could inspire future research. First, our method currently extracts probabilities of part boundaries over points. Sometimes these probabilities seem too low (Figure \ref{fig:part_boundaries_failure_cases}), resulting in sparsely labeled boundary points, which makes it harder to extract a continuous boundary curve. It would be interesting to investigate robust fitting of parametric curves or lines to probabilistic boundaries to localize them more accurately. This could in turn be combined with neural patch fitting \cite{sharma2020parsenet}, and also result in geometric decomposition of point clouds to primitives with more accurately trimmed boundaries. For semantic segmentation, jointly optimizing the unary and pairwise term with the rest of the network in an end-to-end manner could further improve results. Finally, it would be interesting to extend our method to handle polygon mesh segmentations based on our detected boundaries.

\section{Acknowledgements.} 
This research is funded by the
European Union's Horizon 2020 research and innovation programme under grant agreement No 739578 (RISE – Call: H2020-WIDESPREAD-01-2016-2017-TeamingPhase2), 
the Government of the Republic of Cyprus through the Directorate General for European Programmes, Coordination and Development,
and NSF (CHS-1617333). Our experiments were performed in the  UMass  GPU  cluster  obtained  under  the  Collaborative  Fund managed  by  the  Massachusetts  Technology  Collaborative. 
We especially thank Gopal Sharma for assisting with the filtering and preparation of the ABC training data. We thank the anonymous reviewers for their feedback. 

\bibliographystyle{eg-alpha-doi} 
\bibliography{main}       

\newpage

\end{document}